\documentclass[10pt,twocolumn,letterpaper]{article}

\usepackage{iccv}
\usepackage{times}
\usepackage{epsfig}
\usepackage{graphicx}
\usepackage{amsmath}
\usepackage{amssymb}

\usepackage{tikz}
\usepackage{subcaption}
\usepackage{comment}
\usepackage{tabularx}
\usepackage{multirow}
\usepackage{multicol}
\usepackage{booktabs}
\usepackage{enumitem}
\usepackage{bbm}
\newcolumntype{L}{>{\raggedright\arraybackslash}X}
\newcolumntype{C}{>{\centering}p{2cm}}
\definecolor{tgrasp}{RGB}{255, 0, 0}
\definecolor{tcontain}{RGB}{0, 255, 0}
\definecolor{tblue}{RGB}{0, 0, 255}
\usepackage{wasysym}
\renewcommand{\arraystretch}{0.9}

\newcommand{\ra}[1]{\renewcommand{\arraystretch}{#1}}

\DeclareMathOperator*{\argmax}{argmax}

\usepackage{cite}
\usepackage{url}

\usepackage[pagebackref=true,breaklinks=true,letterpaper=true,colorlinks,bookmarks=false]{hyperref}
\hypersetup{
    linkcolor=red,
    filecolor=black,
    urlcolor=gray,
    citecolor=black,
}

\iccvfinalcopy % *** Uncomment this line for the final submission

\def\iccvPaperID{22} % *** Enter the ICCV Paper ID here

\ificcvfinal\fi

\begin{document}

%%%%%%%%% TITLE
\title{Affordance segmentation of hand-occluded containers from exocentric images}

\author{Tommaso Apicella\textsuperscript{1,2}, Alessio Xompero\textsuperscript{2}, Edoardo Ragusa\textsuperscript{1},\\
Riccardo Berta\textsuperscript{1}, Andrea Cavallaro\textsuperscript{2,3,4}, Paolo Gastaldo\textsuperscript{1} \\
%University of Genoa\\
% \and
% Alessio Xompero\\
% \and
% Edoardo Ragusa
% \and
% Riccardo Berta
% \and
%Andrea Cavallaro
% \and
%Paolo Gastaldo
\textsuperscript{1}University of Genoa, Italy, 
\textsuperscript{2}Queen Mary University of London, U.K. \\
\textsuperscript{3}Idiap Research Institute, 
\textsuperscript{4}École Polytechnique Fédérale de Lausanne, Switzerland \\
%{\tt\small tommaso.apicella@edu.unige.it}
{\tt\small \{t.apicella, a.xompero, a.cavallaro\}@qmul.ac.uk}\\
{\tt\small \{edoardo.ragusa, riccardo.berta, paolo.gastaldo\}@unige.it}
}

\maketitle
\ificcvfinal\thispagestyle{empty}\fi

%%%%%%%%% ABSTRACT
\begin{abstract}
Visual affordance segmentation identifies the surfaces of an object an agent can interact with. Common challenges for the identification of affordances are the variety of the geometry and physical properties of these surfaces as well as occlusions. In this paper, we focus on occlusions of an object that is hand-held by a person manipulating it. To address this challenge, we propose an affordance segmentation model that uses auxiliary branches to process the object and hand regions separately. The proposed model learns affordance features under hand-occlusion by weighting the feature map through hand and object segmentation. To train the model, we annotated the visual affordances of an existing dataset with mixed-reality images of hand-held containers in third-person (exocentric) images. Experiments on both real and mixed-reality images show that our model achieves better affordance segmentation and generalisation than existing models. % Data, code, and trained models are available at:~\url{https://apicis.github.io/projects/acanet.html}.
\end{abstract}

%%%%%%%%% BODY TEXT
\section{Introduction}
The term \textit{affordance} refers to the potential interactions between an agent and the environment~\cite{gibson1966senses}. Visual affordance segmentation, the task of identifying affordances in a scene observed by a camera~\cite{hassanin2021visual}, enables assistive technologies for robotics and prosthetic applications (e.g., grasping, object manipulation) or collaborative human-robot scenarios (e.g., handovers)~\cite{christensen2022learning, rosenberger2020object, yang2021reactive}. In this scenario, examples of object affordances include  \textit{grasp} and \textit{contain} for a cup and \textit{grasp} and \textit{cut} for a knife.

Visual affordance segmentation is challenging due to the variety of objects with different affordances; objects of the same category that can vary in their physical and appearance properties; occlusions by other objects that can be present in the image; lighting conditions that can affect the object appearance in the image; and affordance regions in the images that are not always easily identifiable by edges.
Most of the methods for visual affordance segmentation modify existing models for semantic or instance segmentation to predict the affordances of objects placed on a table top~\cite{gu2021visual, myers2015affordance, nguyen2016detecting, nguyen2017object, ragusa2021hardware, rezapour2019towards, yin2022new}. Visual affordance segmentation is even more challenging when the object is hand-held by a person due to the occlusions caused by the hand and the different poses that the object can take. 
Only one method addresses this scenario, but the model does not explicitly consider the presence of the forearm and the hand~\cite{hussain2020fpha}. This can result in inaccurate affordance segmentation. Moreover, the focus of the method on egocentric images from human perspective could be unsuitable for an assistive application, e.g. human-robot collaboration, or generalise to exocentric images.
\begin{figure}[t!]
    \centering
    \includegraphics[width=\linewidth]{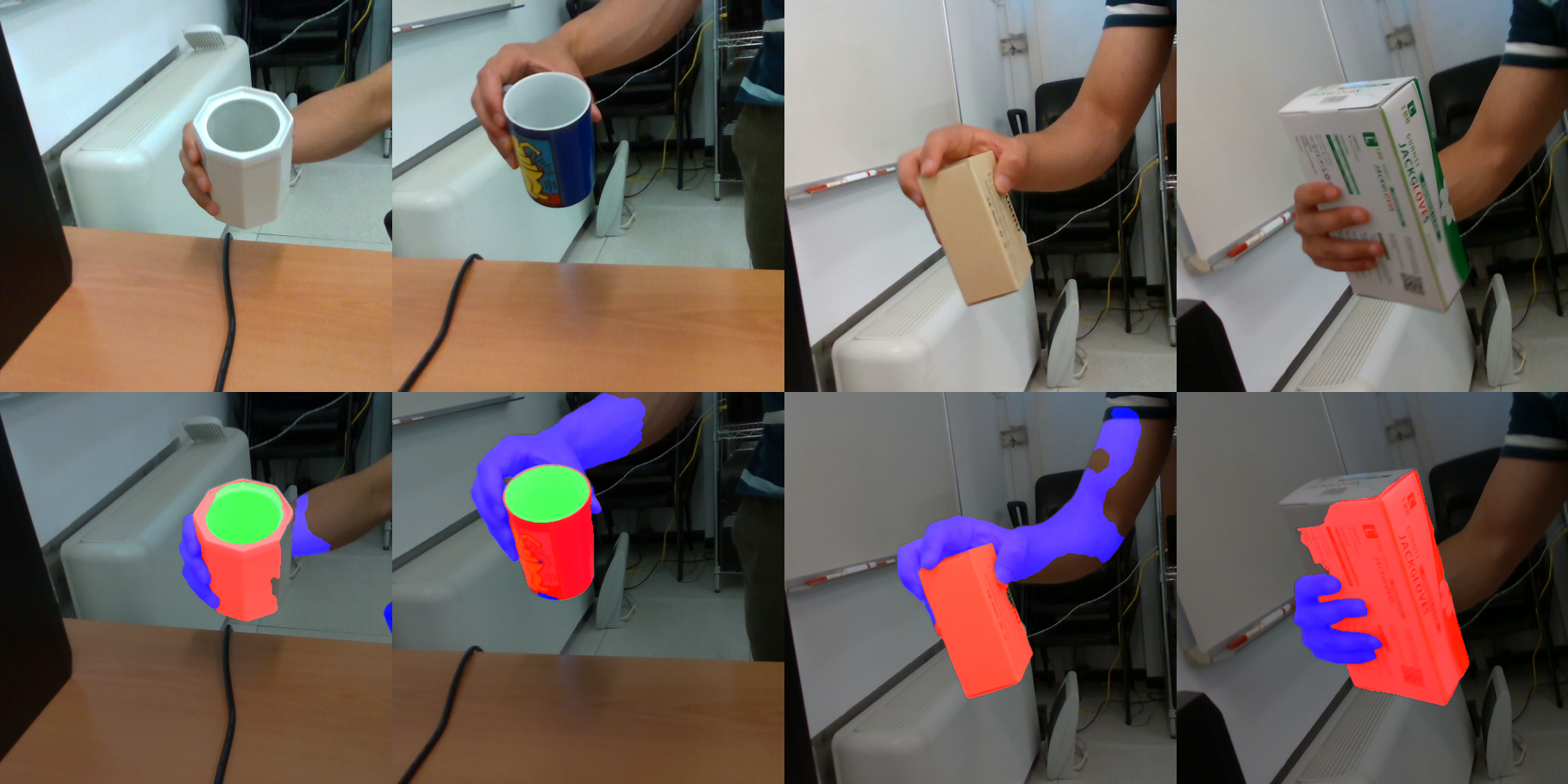}
    \caption{Predictions of the proposed visual affordance segmentation model on RGB images of hand-occluded containers. These images (background and object instances) are never seen during training and do not belong to any of the datasets. Key: 
    \protect\raisebox{2pt}{
    \protect\tikz \protect\draw[tgrasp,line width=2] (0,0) -- (0.3,0);}~\textit{graspable},
    \protect\raisebox{2pt}{\protect\tikz \protect\draw[tcontain,line width=2] (0,0) -- (0.3,0);}~\textit{contain},
    \protect\raisebox{2pt}{\protect\tikz \protect\draw[blue,line width=2] (0,0) -- (0.3,0);}~\textit{arm}. 
    % \textit{Arm} is color-coded as blue to be visible in the overlay.
    % \protect\raisebox{2pt}{\protect\tikz \protect\draw[black,line width=2] (0,0) -- (0.3,0);}~\textit{background},
    }
    \label{fig:vas}
    \vspace{-4pt}
\end{figure}

In this paper, we propose a method that focuses the feature learning on the object and hand region separately, and predicts affordances of an object despite hand occlusions (see Fig.~\ref{fig:vas}). We devise a UNet-like~\cite{ronneberger2015u} multi-branch architecture to predict object and hand segmentation, and we develop a fusion module to learn separate sets of features in the hand and object region.
To train our model, we extend the annotation of an existing mixed-reality dataset of hand-held containers seen from a third-person view with visual affordances. To evaluate the generalisation performance in real settings, we select and manually annotate a set of images from two public datasets with hand-object interaction: HO-3D~\cite{hampali2020honnotate} and CORSMAL Containers Manipulation (CCM)~\cite{xompero2020corsmal}.
Moreover, we re-train existing methods on the annotated dataset and perform a comparative analysis, showing that our proposed model can achieve more accurate segmentation and better generalisation\footnote{Data, code, and trained models are available at \\ \url{https://apicis.github.io/projects/acanet.html}}.

\section{Related work}
\label{sec:relatedwork}
%\begin{comment}
In this section, we discuss existing methods for visual affordance segmentation from RGB images. We also compare existing datasets for this task and discuss their limitations.
%\end{comment}

\subsection{Methods}

Methods for visual affordance segmentation are usually based on models for semantic or instance segmentation, but considering affordance classes as semantic labels~\cite{nguyen2016detecting, ragusa2021hardware, zhang2022multi}. 
% Existing methods can be categorised in a two-stage approach or a direct approach. 
Some of these methods can be part of a two-stage approach that includes a detection step to locate objects of interest. Affordances are then predicted in the region cropped around the detected object or corresponding feature map~\cite{apicella2021affordance,  christensen2022learning, chu2019learning, do2018affordancenet, hassanin2021visual, nguyen2017object, yin2022object}. The two steps can be tackled independently~\cite{apicella2021affordance, nguyen2017object}, to extract the regions of interest, or learned in an end-to-end manner using multi-tasking~\cite{christensen2022learning,
do2018affordancenet, yin2022object}.
For example, AffordanceNet~\cite{christensen2022learning, do2018affordancenet} 
replaces the instance segmentation branch of Mask R-CNN~\cite{he2017mask} to predict affordance classes of the regions of interest localised in the backbone. 
The dependence on the detection step can result in wrongly predicting the presence of an object or missing to localise the object, especially when the object is occluded by a hand. 

Other methods use attention mechanisms to focus on the object area or select more relevant features for the affordance segmentation directly from the input image~\cite{gu2021visual,yin2022new,zhang2022multi, zhao2020object}. 
For example, DRNAtt~\cite{gu2021visual} uses Spatial Attention Module to model contextual information (similarity between features in each pixel position) and Channel Attention Module to model channel inter-dependencies (similarity among channels)~\cite{fu2019dual}. However, DRNAtt could be sensitive to large portions of the image belonging to the background. 
Shared Gradient Attention~\cite{yin2022new} combines affordance segmentation with semantic edge detection. However, this model could be sensitive to edges or object transparencies.

All previous methods focus on scenes with objects that are placed on a tabletop, and are either fully visible or partially occluded due to clutter~\cite{gu2021visual, nguyen2016detecting, yin2022new, zhang2022multi, zhao2020object}. These objects are often opaque or textured and easily distinguishable from the background, % and their placement on the table 
causing the affordance segmentation models to fail when observed objects are in more challenging poses or occluded, as in the case of hand-held objects when manipulated by a person.

There is only one close work that tackles the segmentation of affordances of hand-occluded objects~\cite{hussain2020fpha}. 
This work uses a ResNet-FastFCN~\cite{wu2019fastfcn} with a pyramid parsing module to predict high-resolution outputs by using global contextual information (usually beneficial for segmenting general scenes). However, the model ignores the hand and the global contextual information does not help the model to capture fine affordance predictions on the object of interest. This can result in inaccurate segmentation of the affordances (e.g, predicted on the hand region). 

\begin{figure*}[t!]
    \centering
    \includegraphics[width=\linewidth]{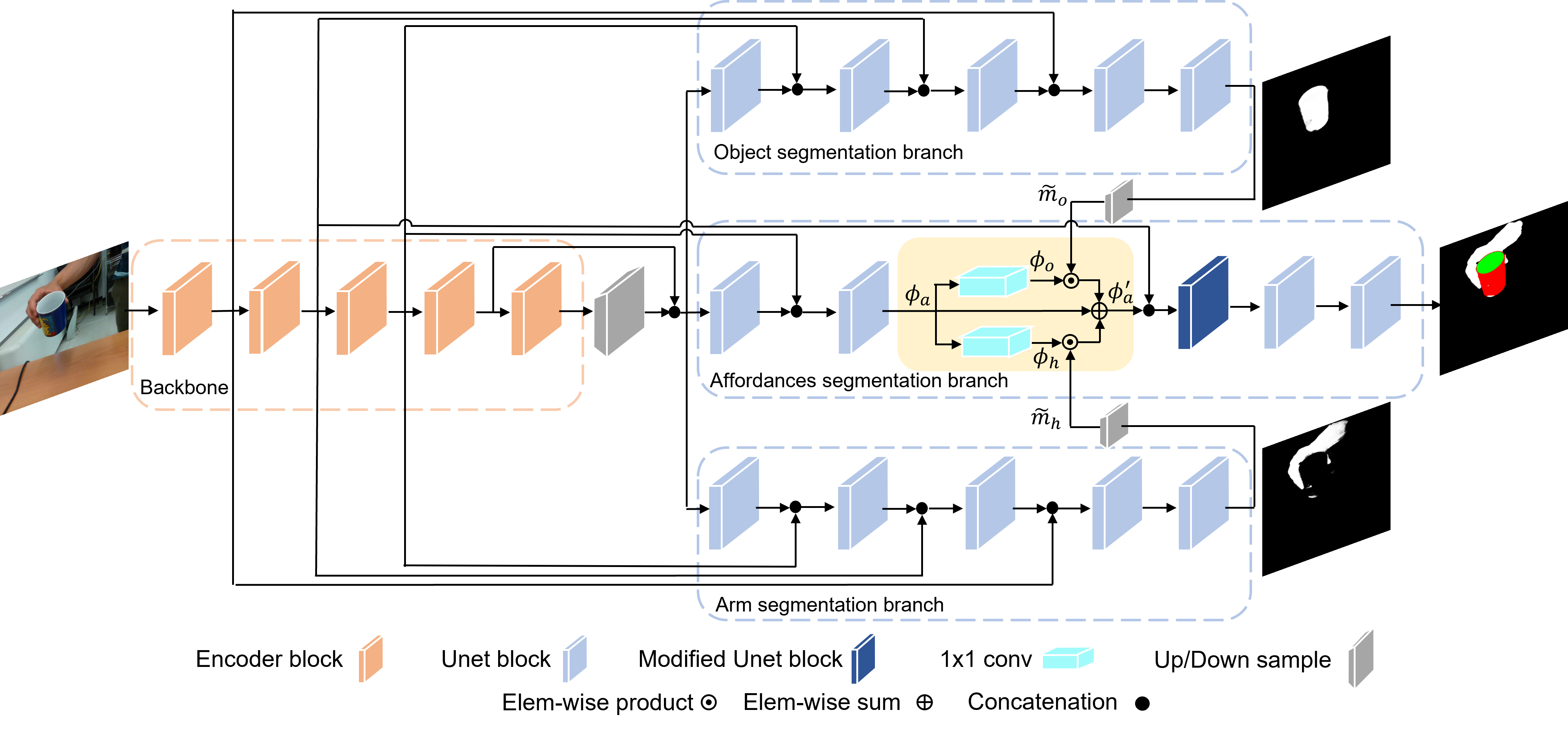}
    \caption{ACANet, our proposed model for Arm-Container Affordance segmentation of hand-held containers. The fusion block is highlighted in yellow.
    }
    \label{fig:proposed_model}
    \vspace{-10pt}
\end{figure*}

\subsection{Datasets} 
\label{ssec:datasets}

Existing datasets are designed to segment the functional (with respect to the design) regions of objects  placed on a table~\cite{christensen2022learning, chu2019learning, jiang2022a4t, khalifa2022towards, myers2015affordance, nguyen2017object, sawatzky2017weakly}. UMD~\cite{myers2015affordance} has 28,843 images of objects placed on a blue rotating table in the same environment, and annotated with  different affordance classes for each object category. Similarly, Multi-View~\cite{khalifa2022towards} has 47,210 images of objects placed on a white rotating table in the same environment, but includes a larger number of affordance classes and object categories. TRANS-AFF~\cite{jiang2022a4t} has 1,346 images of transparent objects on a table, increasing the difficulty to predict affordances. IIT-AFF~\cite{nguyen2017object} has 8,835 images of objects placed in a cluttered scene to better reflect a scenario with occlusions. 
CAD120-AFF~\cite{sawatzky2017weakly} has 3,090 images, sampled from videos of humans performing activities in a realistic setting, e.g., kitchen, office.  
FPHA-AFF~\cite{hussain2020fpha} has 4,300 images of hand-held objects acquired from an egocentric point of view. However, this dataset is currently not publicly available and egocentric images contains arms from the bottom of the image, resulting in objects highly occluded by the hands.

All the previous datasets are manually annotated but limited in their size, making the training of visual affordance segmentation model a problem. The need for thousands of labelled data has pushed towards the generation of synthetic data. UMD-Synth~\cite{chu2019learning} simulates UMD in a synthetic manner and has 37,200 images of objects in different backgrounds and poses, but provides only one affordance per object region. 
AFF-Synth~\cite{christensen2022learning} has 30,245 images generated by using domain randomization to overcome the gap between simulated and real data. Unlike UMD-Synth, each image of AFF-Synth includes multiple objects. These synthetic datasets, however, are still small in size and do not contain occlusions, especially in the case of objects held by a person. Even though CAD120-AFF and FPHA-AFF address the scenario of hand-held objects, a suitable large-size dataset (synthetic or real) for segmenting affordances of objects with different properties (e.g., transparency, shape, size) and under hand-occlusions is missing.

\section{The Arm-Container Affordance Network}
\label{sec:proposed_method}

In this section, we present our multi-branch architecture and the fusion module to perform affordance segmentation (see Fig.~\ref{fig:proposed_model}). We name the model Arm-Container Affordance Network (ACANet) as we consider containers for food and drinks, and \textit{graspable} and \textit{contain} as affordance classes. Assuming the input image has a correctly detected object of interest at the center, the model identifies the classes \textit{graspable}, \textit{contain}, \textit{arm}, and \textit{background}. % as input the image with at the center a correctly detected object of interest, the model identifies the classes \textit{graspable}, \textit{contain}, \textit{arm}, and \textit{background}.  

% Problem formulation
Let $I \in \mathbb{R}^{W\times H \times 3}$ be an exocentric RGB image, where $W$ and $H$ are the image width and height respectively, of a person holding an object, $o$. Let $S \in \{0, ..., C-1\}^{W\times H}$ be a segmentation map where each pixel is one of the $C$ classes that include \textit{background}, \textit{arm}, and the affordances on the visible surface of the object $o$. The objective is to design a model $f(I)$ that assigns each pixel in $I$ to one of the corresponding $C$ classes and that  separates the arm from the object, while identifying the graspable regions of interest on the object from the regions to avoid.

\subsection{Multi-branch architecture}

We devise our multi-branch architecture starting from a UNet-like architecture~\cite{ronneberger2015u} that uses skip connections between the encoder and the decoder to preserve the information and help the gradient flow during back-propagation. We replace the encoder with a ResNet-18~\cite{he2016deep} to include residual connections within the convolutional layers of the encoder, easing the optimization problem~\cite{he2016deep}. Moreover, ResNet keeps the first pooling layer and replaces the others using a stride $2$ in convolutional layers, unlike the original UNet encoder that halves the resolution of tensors using only pooling layers. To double the resolution in the decoders, we replace the UNet convolutional layers with non-trainable up-sampling layers based on nearest interpolation. For each decoder, we also modified the last convolutional layer from $1 \times 1$ kernel and no padding (original UNet) to $3 \times 3$ kernel, stride $1$, and padding. This allows the last convolution to consider neighboring pixels information.

This architecture outputs a tensor $\boldsymbol{S} \in[0, 1]^{W\times H \times C}$ where each channel predicts one of the classes independently, and each pixel in a channel map is a probability such that $\sum_{c=0}^{C-1} S_c(i,j) = 1$, with $i \in \{1,.., W\}$ and $j \in \{1,..., H\}$ being the width and height indices, respectively. 
Including the class \textit{arm} in $\boldsymbol{S}$ already handles the affordance segmentation of hand-occluded containers. However, we experimentally observed that the model first learns the classes with higher number of pixels in the annotation, affecting the prediction accuracy of the other classes. 

We include two additional decoder branches that specialise in the segmentation of the arm and  of the visible region of the object. Segmenting the object helps the model learn the area of the image where the affordances are. For simplicity, we refer to the three decoder branches as Arm, Object, and Affordance segmentation. The \textit{Arm segmentation} branch predicts a probability map, $m_h \in [0, 1]^{W\times H}$, that separates the region associated to the arm (composition of forearm and hand) from all the rest. The \textit{Object segmentation} branch predicts a probability map, $m_o \in [0, 1]^{W\times H}$, that separates the region associated to the visible object (held by the person hand) from all the rest. The \textit{Affordance segmentation} branch fuses the feature maps with the arm and object maps, and predicts the segmentation tensor $\boldsymbol{S}$.

\subsection{Feature separation and fusion}

Object and arm segmentation alone are insufficient to improve the segmentation accuracy under hand-occlusions. We therefore design a module that merges intermediate feature maps $\phi_a \in \mathbb{R}^{C' \times W' \times H'}$, with the down-sampled masks extracted by the \textit{Arm segmentation} branch (\mbox{$\tilde{m}_h \in [0, 1]^{W'\times H'}$}) and the \textit{Object segmentation} branch (\mbox{$\tilde{m}_o \in [0, 1]^{W'\times H'}$}), respectively. We compute the intermediate feature maps within the \textit{Affordance segmentation} branch by using two UNet blocks that process the feature maps outputted by the backbone. The object and arm masks are down-sampled using bi-linear interpolation to match the size of the intermediate feature maps.

Instead of directly combining the features maps with the segmentation masks, we first learn specialised features in the object and arm regions. Specifically, we convolve $\phi_a$ with a set of $1 \times 1$ filters to obtain the feature map related to the object, $\phi_o \in \mathbb{R}^{C' \times W'\times H'}$ (where $C'$ is the number of filters), and with another set of $1 \times 1$ filters to obtain the feature map related to the hand, $\phi_h \in \mathbb{R}^{C' \times W'\times H'}$.
% by ??combining??? the channels of each pixel position of $\phi_a$ with their corresponding set of $1 \times 1$ convolutional filters (the number of filters is $C'$). 
% we project $\phi_a$ into two different feature spaces maintaining the same feature maps resolution, $\phi_o \in \mathbb{R}^{C' \times W'\times H'}$ and $\phi_h \in \mathbb{R}^{C' \times W'\times H'}$, by using $C'$ convolutional filters $1 \times 1$ to combine each pixel position independently.
We then perform a pixel-wise weighting of the feature maps $\phi_o$ and $\phi_h$ with the corresponding segmentation mask $\tilde{m}_o$ and $\tilde{m}_h$. This highly penalises the features outside of the predicted object (or arm) region. The merged feature maps are aggregated with the initial intermediate features as
\begin{equation}\label{eq:feat_separation}
    \phi'_a = \phi_a + (\phi_h \odot \tilde{m}_h) + (\phi_o \odot \tilde{m}_o),
\end{equation}
where $\odot$ is the Hadamard product, i.e., the element-wise product between each feature map in $\phi_o$, $\phi_h$ and the down-sampled segmentation masks $\tilde{m}_o$, $\tilde{m}_h$\footnote{Mathematical simplification as $\tilde{m}_o$ and $\tilde{m}_h$ should be repeated for each channel of the feature maps $\phi_o$ and $\phi_h$.}.

\subsection{Predicting object affordances and the hand}

The \textit{Affordance segmentation} branch uses the fused feature maps, $\phi'_a$, as input to three UNet blocks to predict the output segmentation tensor $\boldsymbol{S}$. Note that the feature maps $\phi'_a$ are concatenated via skip connection with the corresponding intermediate feature maps in the encoder before the first UNet block (see Fig.~\ref{fig:proposed_model}).
We modify the first UNet block after the fusion module to improve the processing of the feature maps. Specifically, we increase the number of output channels of the first convolutional filter from 64 to 128, and we add another convolutional layer to decrease the channels from 128 to 64.
Furthermore, we avoid concatenating low-level information (e.g., edges) from the backbone to the last three UNet blocks via skip connections. This design choice helps the model preserve the changes in the feature maps $\phi'_a$ and predict affordances in the object region. The final segmentation map is:
$S = \argmax_c \boldsymbol{S}$.

% Based on all the architectural design choices, ACANet has $20.82$ million parameters and performs $85.17$ Giga Floating-point operations (GFLOPs).

\subsection{Loss functions}

To train ACANet, we use a linear combination of a Dice loss~\cite{milletari2016v} and two binary cross-entropy losses as:
\begin{equation}
    \mathcal{L} = \: \mathcal{L}_a + \lambda_o \: \mathcal{L}_o + \lambda_h \: \mathcal{L}_h ,
\end{equation}
where the Dice loss, $\mathcal{L}_a$, operates on the affordance branch outputs, the binary cross-entropy, $\mathcal{L}_o$, operates on the object branch output, and the binary cross-entropy, $\mathcal{L}_h$, operates on the arm branch output. The hyper-parameters $\lambda_o$ and  $\lambda_h \in \mathbb{R}$ control the impact of object and hand segmentation losses, respectively. $\mathcal{L}$ allows each branch to specialize for their segmentation task and influences the backbone to learn a common representation for all the branches.
% where the Dice loss, $\mathcal{L}_a = \mathcal{L}_D(y_a, \hat{y}_a)$, operates on the affordance outputs, the binary cross-entropy, $\mathcal{L}_o = \mathcal{L}_{bce}(y_o, \hat{y}_o)$, operates on the object output, and the binary cross-entropy, $\mathcal{L}_h = \mathcal{L}_{bce}(y_h, \hat{y}_h)$, operates on the hand output. The hyper-parameters $\lambda_o$ and  $\lambda_h \in \mathbb{R}$ control the impact of object and hand segmentation losses, respectively. $\mathcal{L}$ allows each branch to specialize for their segmentation task and influences the backbone to learn a common representation for all the branches.

The Dice loss, $\mathcal{L}_a$, addresses the imbalance problem between classes, as the majority of pixels can be labelled as \textit{background}~\cite{sudre2017generalised}. 
Given a batch of $B$ predicted segmentation and corresponding annotations, the Dice loss is\footnote{The margin $\epsilon = 10^{-7}$ avoids numerical issues when  $\boldsymbol{\hat{y}}=\boldsymbol{y}=\boldsymbol{0}$.}:
\begin{equation}
    \mathcal{L}_a =  1 - \frac{1}{C} \sum_{c=0}^{C-1} \frac{2\sum_{b = 1}^{B}  \sum_{l = 1}^{WH} y^c_{l,b} \hat{y}^c_{l,b}}{\epsilon + \sum_{n = 1}^{B}  \sum_{l = 1}^{WH} \hat{y}^c_{l,b} + y^c_{l,b}},
\end{equation}
where $\hat{y} \in [0, 1]^{WH \times C}$ and $y \in \{0, 1\}^{WH \times C}$ are the reshaped predictions and annotations,  respectively, with $\sum_{c=0}^{C-1} y_i^c = 1$. 

The binary cross-entropy loss is used in binary classification and semantic segmentation task, considering each pixel as independent from the others. Given a batch of $B$ predicted object segmentation masks and corresponding annotations, the binary cross-entropy loss for the object is:
\begin{equation}
    \mathcal{L}_{o} =  - \frac{1}{B} \sum_{b = 1}^{B}\sum_{l = 1}^{WH} \boldsymbol{v}_{l,b} \log(\boldsymbol{\hat{v}}_{l,b}) + (1-\boldsymbol{v}_{l,b}) \log(1-\boldsymbol{\hat{v}}_{l,b}),
\end{equation}
where $\boldsymbol{\hat{v}} \in [0, 1]^{WH}$ is the reshaped vector of $m_o$ and $\boldsymbol{v} \in \{0, 1\}^{WH}$ is the reshaped vector of the corresponding annotation. The binary cross-entropy for the hand, $\mathcal{L}_{h}$, is similarly computed using the reshaped vector of $m_h$ and corresponding reshaped annotation.
 
% $\mathcal{L}_{h}$ has the same mathematical formulation, but $\boldsymbol{\hat{v}}$ is the reshape of $m_h$ and $\boldsymbol{v}$ is the arm annotation vector.

\subsection{Training with mixed-reality images}
\label{sec:choc_aff}

Training ACANet requires a large dataset with (exocentric) images of hand-occluded objects and segmentation annotation of both arm, object, and affordances. Such a dataset was not available, and collecting and manually annotating a new dataset is challenging, expensive, and time-consuming. We therefore complement an existing dataset, which has mixed-reality images of hand-occluded containers for object pose estimation~\cite{weber2022mixed}, % and complement the dataset 
with visual affordance annotations. Using mixed-reality datasets can easily scale the generation of a larger number of images under different realistic backgrounds. Moreover, some existing works on hand-object reconstruction or object pose estimation achieved good performance when training on mixed-reality datasets despite the domain gap~\cite{hasson2019learning, wang2019normalized}.

\begin{figure}[t!]
    \centering
    \includegraphics[width=0.9\linewidth]{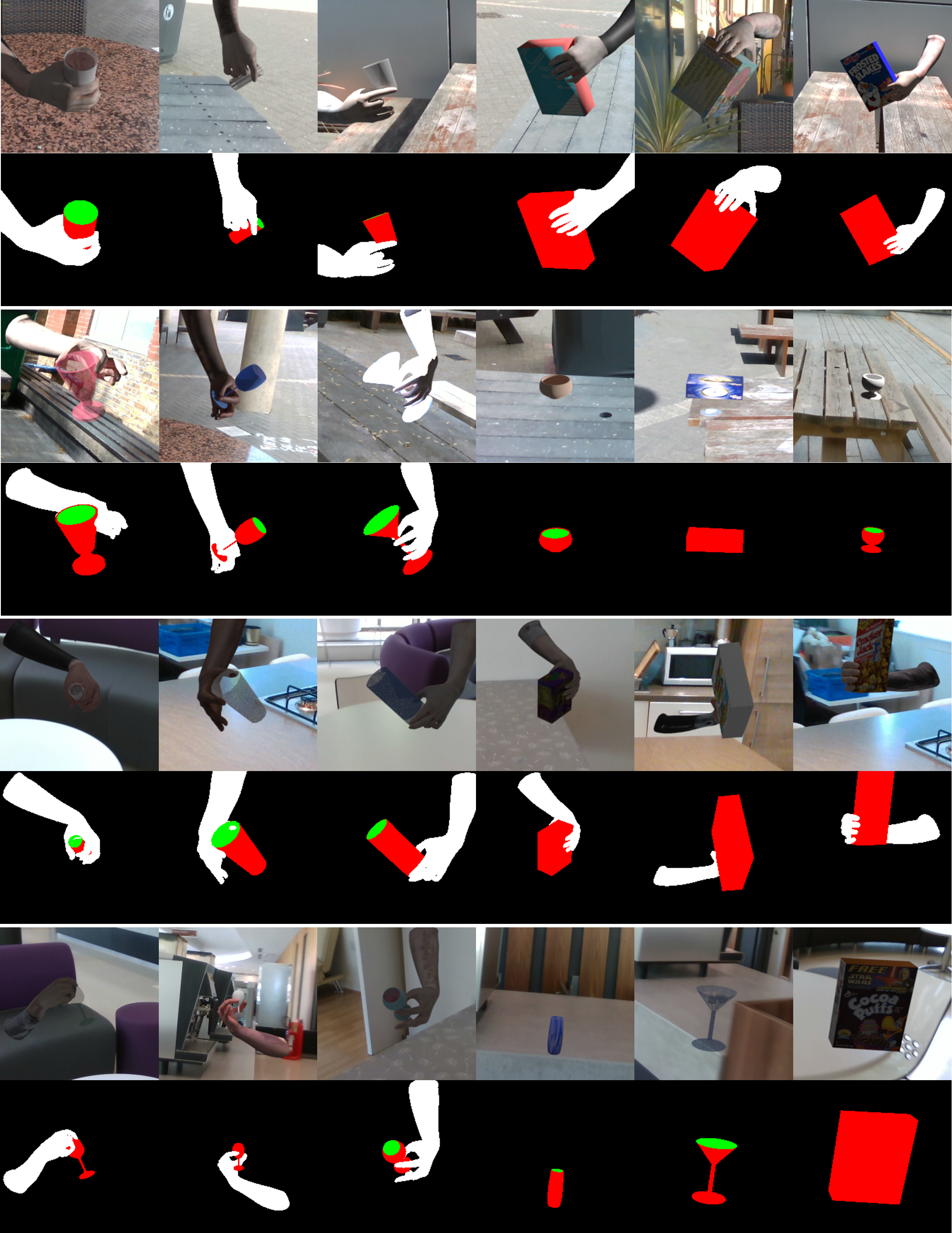}
    \caption{Samples of cropped RGB images and segmentation maps of arms and object affordances from the annotated mixed-reality dataset, CHOC-AFF. Key: 
    \protect\raisebox{2pt}{\protect\tikz \protect\draw[black,line width=2] (0,0) -- (0.3,0);}~\textit{background},
    \protect\raisebox{2pt}{
    \protect\tikz \protect\draw[tgrasp,line width=2] (0,0) -- (0.3,0);}~\textit{graspable},
    \protect\raisebox{2pt}{\protect\tikz \protect\draw[tcontain,line width=2] (0,0) -- (0.3,0);}~\textit{contain},
    \protect\raisebox{2pt}{\protect\tikz \protect\draw[black, thin] (0,0) rectangle (0.25,0.07);}~\textit{arm}
    }
    \label{fig:aff_choc}
    \vspace{-10pt}
\end{figure}

We use the publicly available CORSMAL Hand-Occluded Containers (CHOC) dataset that has 138,240 RGB images of 48 synthetic containers pseudo-realistically rendered on top of 30 different real backgrounds~\cite{weber2022mixed}. The dataset has 8,640 images with objects placed on top of a flat surface and 129,600 images of hand-held objects rendered in various locations and poses above the flat surface in the scene. The 48 containers are evenly distributed among 3 categories (box, stem, non-stem) and vary in their physical properties, such as size and shape, and appearance (textures, transparency). Hand-held objects were generated using synthetic forearms that hold the synthetic containers with three different visually plausible grasps and are orientated towards the pointing direction of the camera to simulate a potential offering of the object. 

These characteristics and the available generation pipeline allows us to easily extend CHOC with the annotation of the \textit{graspable} and \textit{contain} affordances in addition to the existing annotations of the segmentation masks for the \textit{arm} and \textit{object}. We annotate the affordances on the surfaces of the 3D CAD models of the 48 containers using 
Blender\footnote{\url{https://www.blender.org/}}. We then use the object poses annotated in CHOC to project the models with affordances on the image plane and render the affordance maps. We add the hand mask to the affordance map by pixel replacement. For simplicity, we refer to this version of the  dataset with annotations of the affordances as CHOC-AFF. Fig.~\ref{fig:aff_choc} shows sampled images and annotations of CHOC-AFF. We use CHOC-AFF for training and testing ACANet and other models, and we evaluate the generalisation of the models to real images.

\section{Validation}
\label{sec:validation}

\subsection{Methods under comparison}

We compare ACANet against a baseline and two state-of-the-art methods: ResNet18-UNet (RN18-U), ResNet50-FastFCN (RN50-F)~\cite{hussain2020fpha}, and DRNAtt~\cite{gu2021visual}.
RN18-U is the single-branch baseline behind ACANet and has a UNet-like architecture with a ResNet-18 based encoder.  
% RN18-U has $14.32$ million parameters and $38.28$ GFLOPs.
% RN50-F has $66.40$ million parameters and $239.33$ GFLOPs.
DRNAtt is the best performing model on the UMD dataset ~\cite{chen2023survey}. 
We implemented RN18-U and re-implemented DRNAtt to directly segment both the object affordances and the arm. We changed the channels of the last layer in RN50-F\footnote{Original RN50-F implementation~\cite{wu2019fastfcn} is available at~\url{https://github.com/wuhuikai/FastFCN}} to segment only affordances of objects as per the original implementation in case of egocentric data~\cite{hussain2020fpha}.
% We re-implement DRNAtt to segment both the arm and the object affordances. 
% DRNAtt has $17.38$ million parameters and $149.90$ GFLOPs.

\subsection{Experimental setup}
\label{ssec:data_setup}

For our experiments, we split CHOC-AFF into \textit{training} set and \textit{validation} set to train the models, and two \textit{testing} sets to evaluate the models generalisation to different backgrounds and different object instances. We also select and annotate images from two existing public datasets for hand-object pose estimation or reconstruction to evaluate the models in real conditions.

For CHOC-AFF, the training set has $89,856$ images with $26$ out of $30$ backgrounds and $36$ out of $48$ containers ($12$ per object category). The validation set has $17,280$ images with all $30$ backgrounds and $6$ container instances ($2$ per object category) different from the ones in training set. The first testing set has $13,824$ images and evaluates the generalisation performance of the models to the same training object instances in $4$ backgrounds not seen during training. The second testing set has $17,280$ images and evaluates the generalisation performance of the models to $6$ object instances (2 per object category) not seen during training in all $30$ backgrounds~\cite{weber2022mixed}. % This testing set allows us to evaluate the generalisation performance of the models to object instances unseen at training time.

For the two testing sets in real conditions, we consider HO-3D~\cite{hampali2020honnotate} and CCM~\cite{xompero2020corsmal} due to the presence of various challenges, such as presence of the human body, real interactions, and different object instances and hand-object poses. HO-3D is a multi-view video dataset of people manipulating different types of objects. We selected $150$ frames of mugs and boxes as containers from the lateral and frontal cameras (with respect to the arm), keeping a diversity in object and hand poses. We used the segmentation provided by the authors as annotation for the classes \textit{arm}\footnote{Note that the forearm is not annotated.} and \textit{graspable}, whereas we manually annotated the class \textit{contain}. 
CCM is a dataset of multi-view sequences of people manipulating containers with different contents, and then offering the objects to a fixed robot arm. The offering phase allows us to evaluate the models under more realistic human grasps and object poses in a human-robot collaboration scenario, and more challenging conditions caused by different background and lighting settings.  Moreover, containers can vary in their physical appearance (e.g., transparency, texture) or be affected by the presence of content. We selected the last frame (offering phase) of $150$ sequences from a side perspective, diversifying objects, hand poses, and scene color settings. We manually annotated the affordance classes \textit{contain} and \textit{graspable}, and the class \textit{arm} of only the hand(s) in contact with the offered container.

\subsection{Training details and parameters settings} \label{ssec:training_det}

During training, we fine-tune all encoders, pre-trained on ImageNet~\cite{russakovsky2015imagenet}, to start from a better initialisation than random. For ACANet, we set the hyper-parameters $\lambda_o$ and $\lambda_h$ to 1. For all models, we set the batch size to $2$, the initial learning rate to $0.001$, and we use the mini-batch Gradient Descent algorithm as optimizer with a momentum of $0.9$ and a weight decay of $0.0001$. For all models except RN50-F, we schedule the learning rate to decrease by a factor of $0.5$, if there is no increase of the mean Intersection over Union in the validation set for $3$ consecutive epochs. For RN50-F, we set the learning rate schedule following the original setup. We use a Dice Loss to penalise the errors for RN18-U. We use the standard cross-entropy loss for DRNAtt and RN50-F (with auxiliary weight set to $0.2$). We use early stopping with a patience of $10$ epochs to reduce overfitting, and set the maximum number of epochs to $100$. 

Images can be of different resolutions and therefore we apply a cropping square window of fixed size to avoid distorsions or adding padding. Assuming a perfect object detector, we crop a $W \times W$ window around the center of the bounding box obtained from the object mask annotation to restrict the visual field and obtain an object centric view. However, the cropping window can go out of the support of the image if the bounding box is close to the image border. In this case, we extend the side of the window that is inside the image support to avoid padding. 
In case the bounding box is bigger than the cropping window, we crop the image inside the bounding box and resize it to the window size. We apply this cropping procedure to all the images both in training and testing phases.

During training, we also use data augmentation to further increase the diversity of the images.
Specifically, for each input image, we apply the following sequence of transformations: resize by a factor randomly sampled in the interval $[1,1.5]$ to avoid degrading quality; center crop the resized image with a $W \times H$ window to restore the original image resolution; and horizontal flip with a probability of $0.5$ to simulate the other arm. We set the window size to $W=H=480$ (minimum resolution for RN50-F), as higher resolutions degrade the image quality.

\subsection{Performance measures}
\label{subsec:perf_measures}

To evaluate and compare the models, we compute the per-class precision, recall, and Jaccard Index as percentages across all images of a given dataset \mbox{$\mathcal{D} = \{I_n | n=1,...,N\}$}. Precision measures the percentage of true positives among all positive predicted pixels. Recall measures the percentage of true positive pixels with respect to the total number of positive pixels. The Jaccard Index measures how much two regions with the same support are comparable (Intersection over Union or IoU).

To obtain these performance measures, we first compute true positives ($TP$), false positives ($FP$), and false negatives ($FN$) between prediction $S_{n}$ and annotation $G_{n}$ for each RGB image $I_n \in \mathcal{D}$ and for each class $c$. A true positive is a pixel $\boldsymbol{x} \in I_n$ that is predicted as class $c$ in $S_{n}$ and the corresponding pixel in $G_{n}$ is annotated as $c$. A false positive is a pixel $\boldsymbol{x} \in I_n$ that is predicted as class $c$ in $S_{n}$, but not annotated as $c$ in $G_{n}$. A false negative is a pixel $\boldsymbol{x} \in I_n$ that is not predicted as class $c$ in $S_{n}$, but the corresponding pixel in $G_n$ is annotated as $c$.

We therefore compute the per-class precision, $P$, the per-class recall, $R$, and the per-class Jaccard Index, $J$, as:
\begin{equation}
   P = \frac{\sum_{n=1}^{N} \sum_{\boldsymbol{x} \in I_n} TP}{ \sum_{n=1}^{N} \sum_{\boldsymbol{x} \in I_n}  TP + FP},
\end{equation}
\begin{equation}
   R = \frac{\sum_{n=1}^{N} \sum_{\boldsymbol{x} \in I_n} TP}{ \sum_{n=1}^{N} \sum_{\boldsymbol{x} \in I_n} TP + FN},
\end{equation}
\begin{equation}
   J = \frac{\sum_{n=1}^{N} \sum_{\boldsymbol{x} \in I_n} TP}{ \sum_{n=1}^{N} \sum_{\boldsymbol{x} \in I_n} TP + FP + FN}.
\end{equation}
Note that we keep the notation simple and we did not include the index of the class $c$ in the above equations, as all the results will refer to the per-class measures.

\subsection{Results and discussion}
\label{subsec:res_discussion}

Table~\ref{tab:resall} compares the performance of the models on the mixed-reality and real testing sets. For the discussion and ranking of the methods, we consider $J$ as the reference performance measure. 
% Overall comments/takeaways
Overall, ACANet outperforms the other models on all datasets and for most of the classes. This means  that ACANet achieves better generalisation to other backgrounds and object instances in the testing sets.

% Mixed reality testing sets
The performance on the mixed-reality testing sets is similar among the models and $J$ is higher than 90\% for the classes \textit{graspable} and \textit{arm}. Models predict a low number of false positives and false negatives, resulting in a high value for precision and recall ($R$, $P > 95\%$). The class \textit{contain} is the most challenging and $J$ drops to the interval [79\%, 85\%] for the first testing set with unseen backgrounds (CHOC-B) and to the interval [66\%, 70\%] for the second testing set with unseen object instances (CHOC-I). ACANet outperforms the other models for the classes \textit{graspable} and \textit{contain} on both testing sets, whereas DRNAtt has the highest Jaccard Index for the class \textit{arm}. RN50-F has the lowest performance, except for $J$ in \textit{contain} and $P$ in \textit{graspable}. This result shows that adding the \textit{arm} class helps improve the performance. In CHOC-B, models except DRNAtt predict more false positives than negatives for the class \textit{contain} ($P \in [84\%, 91\%]$, $R > 90\%$). On the contrary, the number of false negatives for the class \textit{contain} is higher than false positives ($R \in [71\%, 76\%]$, $P > 88\%$) in CHOC-I.

\begin{table}[t!]
    \scriptsize
    \setlength\tabcolsep{2.1pt}
    \ra{1.1}
    \centering
    \caption{Comparison of the affordance and arm segmentation results between the models on the two mixed-reality testing sets and on the two real testing sets.}
    \begin{tabular}{@{}ll rrr rrr rrr @{}}
    \toprule
    Testing set & {Model} & \multicolumn{3}{c}{\textit{graspable}} & \multicolumn{3}{c}{\textit{contain}} & \multicolumn{3}{c}{\textit{arm}} \\
    \cmidrule(lr){3-5}\cmidrule(lr){6-8}\cmidrule(lr){9-11}
    & & \multicolumn{1}{c}{$P$} & \multicolumn{1}{c}{$R$} & \multicolumn{1}{c}{$J$} & \multicolumn{1}{c}{$P$} & \multicolumn{1}{c}{$R$} & \multicolumn{1}{c}{$J$} & \multicolumn{1}{c}{$P$} & \multicolumn{1}{c}{$R$} & \multicolumn{1}{c}{$J$} \\
    \midrule
    & RN50-F & \textbf{97.33} & 95.72 & 93.27 & 89.83 & 91.94 & 83.27 & - & - & - \\
    & RN18-U & 96.79 & 96.44 & 93.45 & 84.94 & 93.16 & 79.95 & 96.55 & 96.46 & 93.24 \\
    CHOC-B & DRNAtt & 96.38 & \textbf{97.04} & 93.63 & \textbf{91.84} & 90.63 & 83.88 & \textbf{96.94} & \textbf{97.19} & \textbf{94.30} \\
     & \textbf{ACANet} & 97.09 & 96.60 & \textbf{93.88} & 89.46 & \textbf{94.67} & \textbf{85.17} & 96.48 & 96.52 & 93.24 \\
    \midrule
     & RN50-F & \textbf{96.55} & 95.35 & 92.20 & 90.20 & 74.27 & 68.73 & - & - & - \\
     & RN18-U & 96.33 & 96.35 & 92.94 & 88.97 & 74.32 & 68.04 & 96.67 & \textbf{96.91} & 93.78 \\
    CHOC-I & DRNAtt & 95.85 & \textbf{96.74} & 92.85 & \textbf{90.48} & 71.08 & 66.13 & \textbf{97.00} & 96.88 & \textbf{94.07} \\
    & \textbf{ACANet} & 96.36 & 96.51 & \textbf{93.11} & 88.72 & \textbf{76.68} & \textbf{69.86} & 96.94 & 96.77 & 93.90 \\
    \midrule
    \multirow{5}{*}{HO-3D} 
     & RN50-F & \textbf{95.61} & 18.29 & 18.14 & \textbf{90.69} & 79.57 & 73.56 & - & - & - \\
     & RN18-U & 85.85 & 72.53 & 64.79 & 88.21 & 87.61 & \textbf{78.42} & 61.80 & 41.03 & 32.73 \\
     & DRNAtt & 75.42 & 44.08 & 38.54 & 87.26 & 18.75 & 18.25 & 50.23 & 0.32 & 0.32 \\
     & \textbf{ACANet} & 89.72 & \textbf{80.78} &\textbf{73.93}& 79.20 & \textbf{90.43} & 73.07 & \textbf{61.95} & \textbf{53.02} & \textbf{40.00} \\
    \midrule
    \multirow{5}{*}{CCM} 
     & RN50-F & 6.14 & 87.87 & 6.09 & 13.51 & 33.11 & 10.61 & - & - & - \\
     & RN18-U & \textbf{13.69} & 78.69 & \textbf{13.20} & 31.92 & \textbf{42.44} & 22.28 & 44.21 & 42.53 & 27.68 \\
     & DRNAtt & 6.37 & \textbf{95.09} & 6.35 & 0.00 & 0.00 & 0.00 & 4.47 & 0.24 & 0.23 \\
     & \textbf{ACANet} & 10.22 & 86.50 & 10.06 & \textbf{45.40} & 37.46 & \textbf{25.83} & \textbf{49.47} & \textbf{45.35}& \textbf{31.00} \\
    \bottomrule \addlinespace[\belowrulesep]
    \multicolumn{11}{l}{\parbox{0.9\columnwidth}{\scriptsize{Highlighted in \textbf{bold} the proposed model, and the best performing results.}}}\\
    \multicolumn{11}{l}{\parbox{0.98\columnwidth}{\scriptsize{KEY -- RN50-F:~ResNet50-FastFCN~\cite{hussain2020fpha}, RN18-U:~ResNet18-UNET, DRNAtt~\cite{gu2021visual}, J:~per-class Jaccard Index, CHOC-B:~the CHOC-AFF testing set with new backgrounds, CHOC-I:~the CHOC-AFF testing set with new instances.}}}\\
    \end{tabular}
    \label{tab:resall}
    \vspace{-10pt}
\end{table}

\begin{figure*}[t!]
    \centering
    \includegraphics[width=0.96\linewidth]{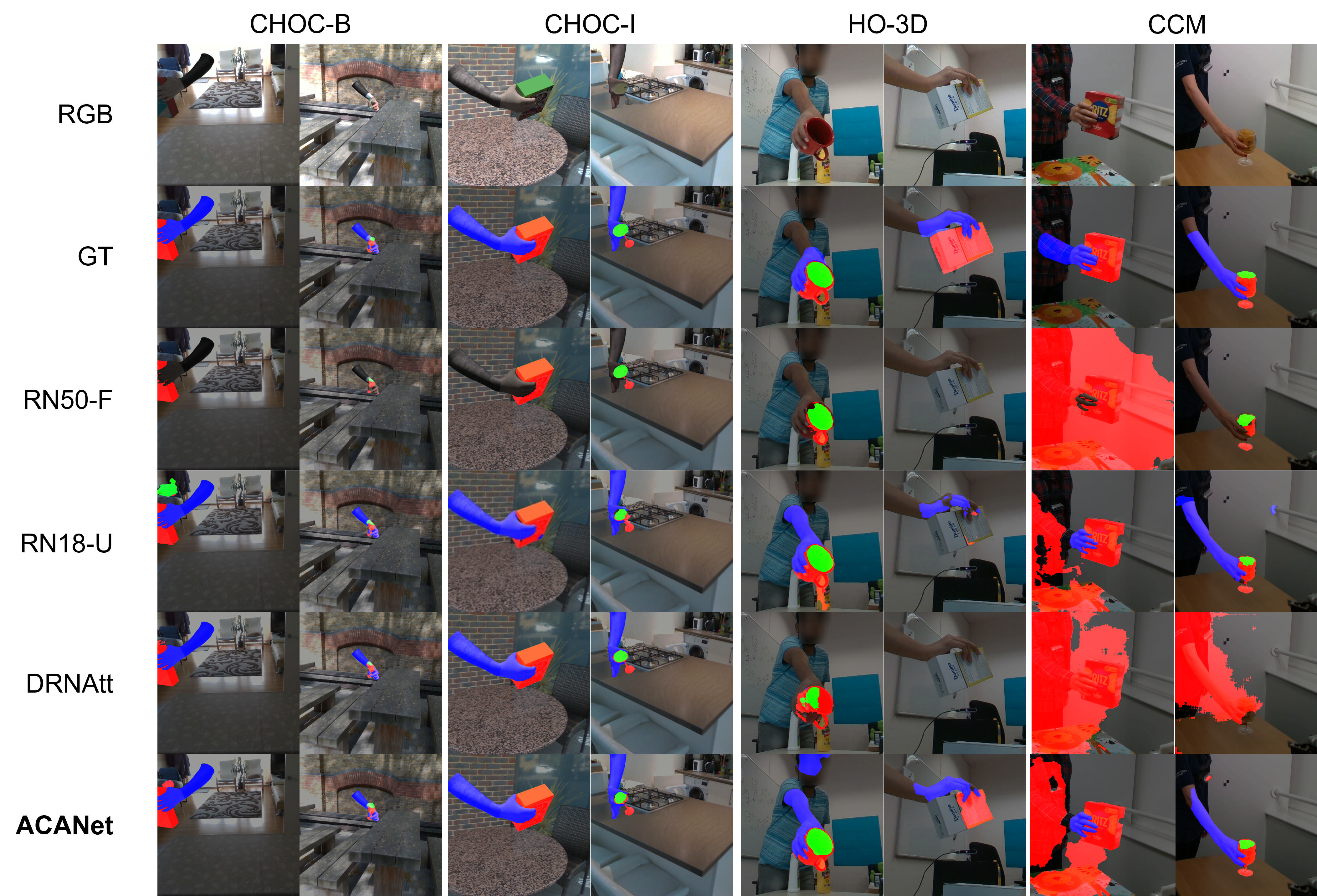}
    \caption{Comparison of the predicted affordance and hand masks of the  models on sampled  images from the four testing sets. The segmentation masks are overlayed on the RGB images. KEY - GT:~ground-truth,
    \protect\raisebox{2pt}{
    \protect\tikz \protect\draw[tgrasp,line width=2] (0,0) -- (0.3,0);}~\textit{graspable},
    \protect\raisebox{2pt}{\protect\tikz \protect\draw[tcontain,line width=2] (0,0) -- (0.3,0);}~\textit{contain},
    \protect\raisebox{2pt}{\protect\tikz \protect\draw[tblue, line width=2] (0,0) -- (0.3,0);}~\textit{arm}. %\textit{Arm} is color-coded as blue to be visible in the overlay.
    }
    \label{fig:qual_res}
    \vspace{-10pt}
\end{figure*}

The HO-3D and CCM testing sets allow us to assess the generalisation capabilities of the models to images acquired in real scenarios, given the known problem of the gap between synthetic and real data. As expected, performance of the models is lower in the real testing sets than in the mixed reality testing sets due to the domain shift. This can be observed especially for the classes \textit{graspable} and \textit{arm}. 

In HO-3D, ACANet outperforms the other models for the classes \textit{graspable} and \textit{arm}. However, all models tend to predict the wrong class in the \textit{graspable} and \textit{arm} regions ($P > R$), and even ACANet has a high  number of false positives and false negatives for the class \textit{arm} ($J = 40\%$).
The performance for the class \textit{arm} is penalised for all models due to the lack of annotation of the forearm, and the presence of the human body and challenging arm poses. 
%For all models, the difference between $R$ and $P$ in the class \textit{graspable} is more than $60\%$ percentage points (p.p.). For \textit{graspable} and \textit{arm}, ACANet has the smallest difference between precision and recall (maximum $9$ p.p.), whereas the other models have larger differences (e.g., $31$ p.p. for DRNAtt). 
For the class \textit{contain}, DRNAtt predicts a high number of false negatives that affect the final performance ($J=18.25\%$), whereas the other models predict a lower number of false positives than DRNAtt, resulting in a higher Jaccard index ($J \in [73.07\%, 78.42\%]$).

In CCM, the tablecloth and the presence of the human body are the main challenges for the models, causing a performance drop compared to the other datasets. In the presence of the tablecloth, models tend to predict \textit{graspable} in most regions of the image. This results in a large difference between $P$ and $R$ (e.g., $76$ percentage points for ACANet). % This result suggests that the models tend to predict \textit{graspable} in most regions of the image, covering also the regions that are actually \textit{graspale}. 
ACANet achieves the best performance for the classes  \textit{contain} ($J=25.83\%$) and  \textit{arm} ($J=31\%$). DRNAtt does not generalise to the real images of CCM with $J \leq 1\%$ for the class \textit{arm} and $J=6.35\%$ for the class \textit{graspable}, whereas the class \textit{contain} is not predicted. This is caused by a large number of false positives towards the class \textit{graspable} ($P=6.37\%$) and a large number of false positives and false negatives for the class \textit{arm} ($P=4.47\%$, $R=0.24\%$). The higher performance of ACANet compared to   DRNAtt and RN18-U shows that learning arm and object features separately is better than learning affordances directly.

Fig.~\ref{fig:qual_res} shows and compares the affordance segmentation results of the models on sample images from the four testing sets. We chose images with objects and hand poses that are challenging and never seen in training, e.g., holding a box from the bottom; and with different  backgrounds and lighting or different object appearances. Visually, ACANet achieves the most accurate segmentation for the arm and the object affordances (see also Fig.~\ref{fig:vas}). For CHOC-B (columns 1 and 2), the predictions have more false positives in the background, whereas there are false positives or false negatives in the object region for CHOC-I (columns 3 and 4). For HO-3D, ACANet predictions show better affordance segmentation than other models, but the number of false positives of the class \textit{arm} increases in presence of the human face (column 5). All models show a high number of false positives for the class \textit{graspable} in the CCM testing set, especially when there is a colorful tablecloth (7th column). However, the predictions of RN18-U and ACANet are close to the annotation when the setting is similar to the training one, i.e., there is no colorful tablecloth nor the human body/face, but just the arm holding the container (columns 6 and 8). In the last column, we also chose to show the results for a transparent cup as transparency can be a challenge for the models due to the not clearly defined borders. For example, the background environment or the content in the container can influence the segmentation predicted by the models. In the sampled image, the cup is filled with a content, and the models are able to correctly predict both the \textit{contain} and \textit{graspable} regions, except for DRNAtt. 

\begin{table}[t!]
    \centering
    \footnotesize
    \caption{Models size and computational cost.
    \vspace{-5pt}
    }
    \begin{tabular}{lrrrr}
    \toprule
    & ACANet & RN18-U & RN50-F & DRNAtt \\
    \midrule
    \# parameters & 20.82 & 14.32 & 66.40 & 17.38 \\
    GFLOPs     & 85.17 & 38.28 & 239.33 & 149.90 \\
    \bottomrule
\addlinespace[\belowrulesep]
    \multicolumn{5}{l}{\parbox{0.9\columnwidth}{\scriptsize{KEY: \# parameters:~number of parameters (in millions), GFLOPs:~Giga Floating-point operations, RN50-F:~ResNet50-FastFCN~\cite{hussain2020fpha}, RN18-U:~ResNet18-UNET, DRNAtt~\cite{gu2021visual}.}}}\\
    \end{tabular}    
    \label{tab:complexity}
    \vspace{-10pt}
\end{table}

We also briefly discuss the complexity of the models measured in number of parameters (in millions) and  number of operations (as Giga Floating-point or GFLOPs).  
% ACANet has $20.82$ million parameters and performs $85.17$ Giga Floating-point operations (GFLOPs). RN18-U has $14.32$ million parameters and $38.28$ GFLOPs. RN50-F has $66.40$ million parameters and $239.33$ GFLOPs.   DRNAtt has $17.38$ million parameters and $149.90$ GFLOPs.
Table~\ref{tab:complexity} shows that RN50-F has the highest amount of parameters and operations, but the generalisation performance is worse than ACANet.
DRNatt has $3$ millions fewer parameters than ACANet, but $1.7 \times$ more operations than ACANet and the performance on real-data is worse in all classes. Finally, ACANet has an increased number of parameters ($6$ millions) and GFLOPs ($2.2$ times) compared to the baseline RN18-U, thus contributing to the higher performance in almost all classes and testing sets.

\section{Conclusion}
\label{sec:conclusion}

We tackled the problem of visual affordance segmentation of hand-occluded containers and proposed ACANet, a multi-branch convolutional neural network that fuses object and hand segmentation mask with the affordance features to specialise filters for object and hand regions.
Training ACANet on an annotated dataset with mixed-reality images of hand-held containers leads to better generalisation to real images with containers not seen in training and with new backgrounds (e.g., on the HO-3D dataset or on-the-fly acquired images). Moreover, ACANet outperforms alternative methods when segmenting the \textit{graspable} area and the person's \textit{arm}, and can achieve a more accurate segmentation even when the objects are in more challenging poses caused by how the person holds the object.

As future work, we will use a larger set of objects, perform an ablation study on the fusion module components, reduce the computational costs necessary to run the model, and validate it in a human-robot collaboration scenario.

{\small
\bibliographystyle{ieee_fullname}
\bibliography{egbib}
}

\end{document}